\documentclass[10pt,twocolumn,letterpaper]{article}

\usepackage{cvpr}              % To produce the CAMERA-READY version
\usepackage{graphicx}
\usepackage{amsmath}
\usepackage{amssymb}
\usepackage{booktabs}
\usepackage{colortbl}
\usepackage{multirow}
\usepackage{makecell}
\usepackage[accsupp]{axessibility} % Improves PDF readability for those with visual impairments.
\usepackage[dvipsnames]{xcolor}

\definecolor{cvprblue}{rgb}{0.21,0.49,0.74}
\usepackage[pagebackref,breaklinks,colorlinks,citecolor=cvprblue]{hyperref}

\def\paperID{5247} % *** Enter the Paper ID here
\def\confName{CVPR}
\def\confYear{2024}

\title{Non-autoregressive Sequence-to-Sequence Vision-Language Models}

\author{Kunyu Shi \quad Qi Dong \quad Luis Goncalves \quad Zhuowen Tu \quad Stefano Soatto\\
AWS AI Labs
\\
{\tt\small \{kunyus, qdon, luisgonc, ztu, soattos\}@amazon.com}
}
\begin{document}
\maketitle

%%%%%%%%% ABSTRACT
\begin{abstract}

Sequence-to-sequence vision-language models are showing promise, but their applicability is limited by their inference latency due to their autoregressive way of generating predictions.
We propose a parallel decoding sequence-to-sequence vision-language model, trained with a Query-CTC loss,
that marginalizes over multiple inference paths in the decoder. 
This allows us to model the joint distribution of tokens, rather than restricting to conditional distribution as in an autoregressive model. The resulting model, NARVL, achieves performance on-par with its state-of-the-art autoregressive counterpart, but is faster at inference time, reducing from the linear complexity associated with the sequential generation of tokens to a paradigm of constant time joint inference. Code is available at: \url{https://github.com/amazon-science/NARVL}.

\end{abstract}

%%%%%%%%% BODY TEXT
\section{Introduction}
\label{sec:intro}

Sequence-to-sequence autoregressive Transformers \cite{vaswani2017attention, floridi2020gpt, raffel2020exploring} are deep neural network architectures that map a sequence of tokens, each representing a segment of text as a vector, onto another sequence, typically representing the same sequence shifted forward by one. Such models can handle a variety of tasks \cite{raffel2020exploring, lewis2019bart, paolini2021structured}, whereby the input (query) text could be a sentence in natural language, and the output (target) the same sentence in a different language (translation), or the answer to a question expressed in the input  (question-answering, QA), the name of an entity or class, etc. The Transformer architecture's versatile and unified design has led to the development of all-in-one (AIO) models, such that multiple tasks can be approached as a sequence-to-sequence translation problem.

\begin{figure}[!t]
%\vspace{-15mm} 
\centering
\includegraphics[width=0.48\textwidth]{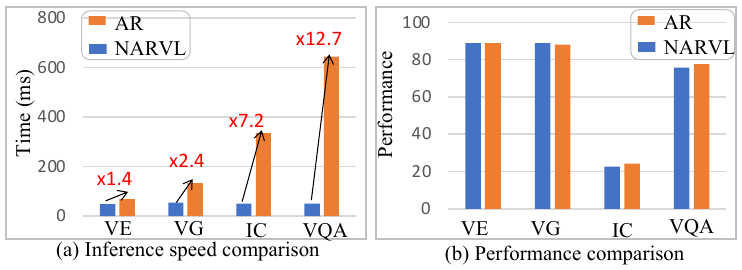}
% \vspace{-5mm}
\caption{Comparison of inference speed and performance between  NARVL (non-autoregressive) and its autoregressive counterpart on four vision-language tasks: Visual entailment (VE), Visual grounding (VE), Visual Question Answering (VQA), and Image captioning (IC). From (a), we see that NARVL speeds up the inference of AR by a factor between 1.4 and 12.7, while achieving on-par performance. 
}
\label{fig:hist}
\end{figure}

Vision-Language AIO Models \cite{yang2021crossing,wang2022ofa,lu2022unified,Gao_2024_CVPR}, including sequence-to-sequence, have proven successful at mapping multimodal inputs, typically images and strings of text, to textual outputs that encode tasks expressible as a string of text, such as visual question answering (VQA), visual grounding (VG), visual entailment (VE), and image captioning (IC). These auto-regressive sequence-to-sequence models face the inference cost issue, since they tend to be unwieldy and need to be executed $T$ times to generate an output sequence of length $T$.

Non-autoregressive methods are proposed in some recent Visual-language AIO models \cite{kamath2021mdetr}, which formulate sequence-to-sequence mapping as a bipartite matching problem. This approach excels in tasks where visual information is key, such as object grounding and detection. However, it's less effective of handling language-focused tasks like Visual Question Answering and Image Captioning. This discrepancy may stem from the nature of the tasks: in object detection/grounding, tokens are orderless and each token correlates to distinct objects or boxes, leading to a weaker inter-object correlation compared to the stronger inter-word correlation in sentences where tokens are ordered. Consequently, the set-to-set, order-independent translation method is more suitable for visual tasks than for language-oriented ones.

\noindent{\bf Main hypothesis:} We hypothesize that a transformer-based architecture could leverage the homogeneity of the input and output spaces, while  enabling more flexible output spaces. In particular, we are interested in the possibility of performing {\em joint decoding} of a sequence in one step, rather than step-by-step. We test whether such an architecture could achieve performance comparable to the auto-regressive baseline at significantly reduced inference cost.  

To test this hypothesis, we develop a new Visual-language AIO model, turning a Transformer-based autoregressive one-step prediction model into a {\em joint predictor} of the target tokens, as explained in Sect.~\ref{sec:method}. In Sect.~\ref{sec:experiments} we show that such a model, which we name NARVL, can be used for the multiple visual-language tasks of interest (VQA, captioning, entailment, grounding). As shown in Fig \ref{fig:hist}, NARVL achieves comparable performance to a state-of-the-art autoregressive model, with significant speed advantage ranging from 1.4 to 12.7 times.

NARVL is made possible by re-purposing of the decoder of an autoregressive Transformer model, and the model has a layer of {\em learnable query tokens} (LQT) that are fixed at inference time and learned during fine-tuning. NARVL is enabled by Query-CTC (Q-CTC) loss, a variant of the CTC loss used in audio and language \cite{graves2006connectionist} but never applied to the visual domain, where the ordinary empirical cross-entropy loss (CE) is marginalized with respect to generative variability in the prediction. Whereas in the language domain the multiple decoding hypotheses stem from the output of the encoder, in vision this is limiting, since input and output spaces are heterogeneous. Therefore, we modify the CTC loss to marginalize not with respect to decoding paths, but with respect to paths from the {\em sequential} learnable query tokens of order indexes to the predicted tokens. 

Our {\bf key contributions} can therefore be summarized as follows: (i) we propose a new sequence-to-sequence {\em non-autoregressive} all-in-one vision language model, that generates sequences in-parallel. (ii) We introduce Query-CTC loss to train this architecture, inspired by the CTC loss used in audio recognition and language, that leverages the sequential learnable query tokens to generate multiple generative paths, and marginalizes the resulting population in the ordinary cross-entropy loss. We show that (iii) the resulting architecture is competitive with state-of-the-art auto-regressive architecture in multiple vision-language tasks, at a significantly reduced inference time, since the model is executed once at inference time, rather than sequentially for as many steps as tokens in the output layer.

\begin{figure*}[!th]
%\vspace{-15mm} 
\begin{center}
\includegraphics[width=1\textwidth]{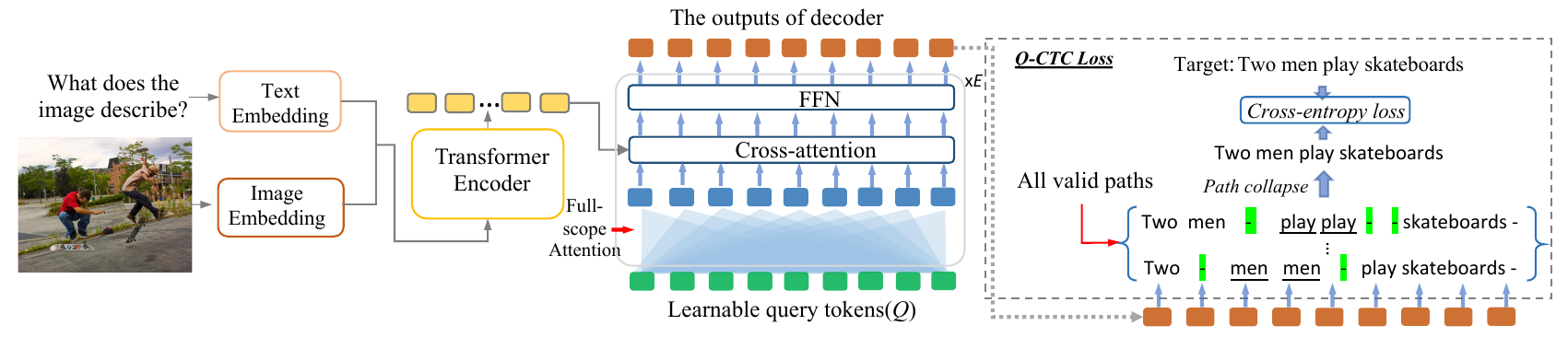}
\end{center}
% \vspace{-5mm}
\caption{\textbf{The overview of NARVL.} NARVL borrows the encoder from OFA \cite{wang2022ofa}, where the embedding sequence of input text and image CNN (ResNet) feature are concatenated in the input token sequence. Unlike the standard transformer decoder that generates outputs sequentially, conditioning on the generated sequence, our non-autoregressive decoder takes a sequence of tokens that are learnable weights, and generates outputs for all tokens in parallel. As the output sequence length is unknown,  we set the number of of learnable query tokens to a value (hyperparameter) larger than the largest target sequence length. The loss used, Q-CTC, is described in Eq.~\ref{eq:LQC}. }
% \vspace{-4mm}
%\vspace{2mm}
\label{fig:overall_architecture}
\end{figure*}

%------------------------------------------------------------------------
\section{Related Work}
\label{sec:formatting}
\noindent{\bf Sequence to Sequence Generation}. Many NLP tasks share a common problem setting where the input consists of  sequences of words with the output being targeted sequences.
Therefore, the sequence to sequence formulation becomes a prototypical setting for many tasks in NLP \cite{sutskever2014sequence}, which can be readily solved by an autoregressive (AR) model.  Beyond recurrent neural networks (RNNs), the decoder in Transformers \cite{vaswani2017attention} also adopts the autoregressive strategy in both training and prediction. On the vision side, considering an image as a sequence of tokens was popularized by the Vision Transformer  (ViT) \cite{dosovitskiy2020image}. Transformers \cite{vaswani2017attention} based approaches pix2seq \cite{chen2021pix2seq, chen2022unified} formulate object detection, instance segmentation, and human poses as sequence generation problem, which differs a lot from approaches designed specifically for individual tasks \cite{carion2020end, lazarow2020learning, Dong_2021_ICCV}. Furthermore, \cite{wang2022ofa, zhang2023musketeer} unify more tasks with seq2seq models. Recently, vision-language tasks have received increasing attention \cite{mao2014deep}, including visual questioning and answering \cite{antol2015vqa}, and visual grounding \cite{yu2016modeling} that have also been tackled by AR models for sequence to sequence generation.

\noindent{\bf Non-autoregressive Sequence Generation}. \cite{gu2017non} proposes non-autoregressive (NAR) Transformer model for machine translation that generates the translated sequence in parallel. The main challenge in NAR is to capture the inter-token dependency \cite{gu2020fully} since the predictions are made conditionally independent. To improve  inter-token dependency, in \cite{sun2019fast,wang2018semi,stern2018blockwise,stern2019insertion,gu2019levenshtein,xiao2022survey}, the model architectures are modified to conduct local NAR only, or light AR layers are added at the end of the decoder. In \cite{ran2021guiding,akoury2019syntactically,shu2020latent,kaiser2018fast,bao2022latent}, hand-crafted or learnable latent variables are incorporated. Knowledge distillation  \cite{zhou2019understanding} has proven effective to reduce complexity of training and increase robustness. In \cite{ghazvininejad2019mask,liu2020task}, curriculum learning has been applied to simplify the learning task.\\
\noindent{\bf Set-to-Set Prediction}. In the seminal Detection Transformers (DETR) \cite{carion2020end}, a set of object queries are turned into a set of detected objects. A key component is the Hungarian matching step that deterministically assigns object queries to the ground-truth objects in training. Broadly speaking, we can also view the decoder in DETR as performing non-autoregressive set generation, which works well for orderless objects but may not be directly applicable to text sequence generation in vision-language models.

\noindent{\bf Non-autogressive Vision-Language Models}. Inspired by \cite{gu2017non}, we develop NARVL, a non-autogressive vision-language model and illustrate the effectiveness of NARVL on a recent autogressive model, OFA \cite{wang2022ofa}. The Connectionist Temporal Classification (CTC) loss \cite{graves2006connectionist} and knowledge distillation \cite{zhou2019understanding} have been adopted in NARVL to implement a NAR solution.  CTC is designed to align the two sequences with different lengths, and it marginalizes over all possible monotonic alignments. It assumes output always longer than target, and the final output is decoded by collapsing repetitive tokens. The variant of the CTC loss we introduce is formalized in Eq.~\ref{eq:LQC}, and NARVL described in Sect.~\ref{sec:method}.

\section{Method}
\label{sec:method}

We co-opt a pre-trained sequence-to-sequence autoregressive model (OFA \cite{wang2022ofa}) and turn it from a one-step predictor (AR) encoding of $p(y_{t+1} | x_1, \dots x_T, y_1, \dots y_t)$ to a flexible task-dependent joint encoding of the target query given heterogeneous inputs
$p(y_1, \dots, y_N | x_1, \dots x_T; q)$, where $x_i$ are token embeddings of images and text and $y$ are output token embeddings. The overview of the proposed NARVL is shown in Figure \ref{fig:overall_architecture}.

Specifically, we embed an image with a convolutional backbone and obtain an activation map with $D$ channels, which we represent as $W\times H$ tokens of dimension $D$, each representing one among the $W\times H$ pixels, along with a positional encoding. 
We concatenate visual tokens with textual tokens, obtained from the input string with BPE tokenization \cite{sennrich2015neural}, to form the input to the encoder, which is identical to \cite{wang2022ofa}. 
The decoder, however, is different from OFA \cite{wang2022ofa}, although it shares the overall structure, thus enabling us to leverage its pre-trained weights to fine-tune. 

\subsection{Parallel Transformer Decoder}
\begin{figure*}[!th]
\centering
\begin{tabular}{ccc}
\includegraphics[width=0.95\linewidth]{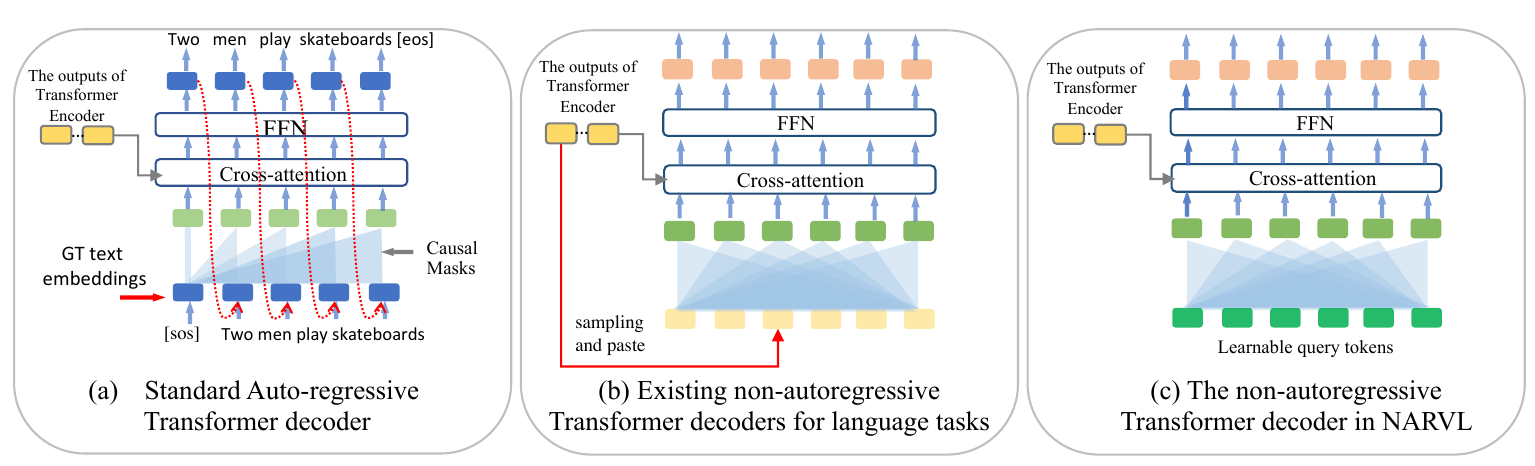}
\end{tabular}
\caption{Comparison of various design of Transformer decoder. (a) Standard Auto-regressive Transformer decoder; (b) The existing non-autoregressive
Transformer decoders for audio and language tasks; (c) The proposed non-autoregressive
Transformer decoder in NARVL. During training, an AR decoder (a) uses teacher forcing with causal masks, where tokens can only attend to previous tokens, while all tokens can attend to each other in the decoders of (b) and (c). NARVL decoder has dedicated query tokens inputs, instead of using the outputs of the encoder as inputs in (b). This design avoids the large latency of the decoder due to the long output sequence from the encoder. }
\label{fig:nar}
\end{figure*}

The OFA decoder is an auto-regressive predictor that takes as input the output of the encoder and a sequence of $T$ consecutive input tokens, % $\{x_{i-K}, \dots x_i\}$ and 
and produces as output the same sequence shifted forward by one, limiting the hypothesis space to the range of the $T+1$ token, and forcing repeated execution $T$ times at inference time, to produce the single last token (see Figure \ref{fig:nar}(a)). 
Our NARVL decoder also takes as input the output of the encoder, but it replaces the input sequence {\em  with a fixed-length and constant sequence of tokens,} representing the task-specific hypothesis space implicit in the training data (see Figure \ref{fig:nar}(c)). 

This constant layer of {\em Learnable Query Tokens} (LQT), is chosen during training as the output of a module with learnable parameters, including the number of tokens, as a hyperparameter, bounded from below by a function of the length of the ground truth output sequence (target tokens) in the training set.

\noindent{\bf{Discussion}} The design of NARVL decoder is different from the existing non-autoregressive Transformer decoder proposed in Natural Language processing \cite{gu2020fully,gu2017non, gu2019levenshtein} (see Figure \ref{fig:nar}(b)), which utilizes the outputs of the encoder as the inputs to the decoder. However, unlike translation, the sequence lengths of input and output sequences are not strongly correlated in vision-language tasks. The NARVL encoder-decoder design is also different from set-to-set decoders \cite{carion2020end, kamath2021mdetr}, and our learnable queries for the decoder is a sequence rather than a set. For each learnable query, we add absolute position embeddings in the bias of attention layers, to force the structure of output sequences.

\subsection{NARVL with Query Connectionist Temporal Classification Loss }
\label{sec:LQC}

NARVL is trained with a variant of the CTC loss \cite{graves2006connectionist} used in audio and language, which consists of a cross-entropy marginalized over a distribution of predictions. In our case, the distribution of prediction corresponds to multiple inference paths in the decoder from the redundant learnable query tokens, to the smaller sequence of output tokens. 

\subsubsection{Q-CTC for Visual-Language Translation }
Suppose we have a training set of $n$ image-text pairs $\mathcal{D}=\{(\mathbf{I}_{i}, \mathbf{T}_{i},\mathbf{Y}_{i} )\}_{i}^{n}$ where $\mathbf{I}$, $\mathbf{T}$, $\mathbf{Y}$ specify the input images, input texts and target outputs, respectively. Note that the output sequences can be text sequences or location sequences. The model generates $\mathbf{Y}$ according to the inputs $\mathbf{I}$ and $\mathbf{T}$.
In order to formalize the expression of the Q-CTC loss,we call the Learnable Query Tokens (LQT) $q \in {\mathbb R}^{D\times N}$, the outputs of the encoder $x \in {\mathbb R}^{D\times L}$, the output tokens are $z \in {\mathbb R}^{D\times N}$, and the ground truth tokens $y \in {\mathbb R}^{D\times T}$, where $N > T $. Let the decoder vocabulary denote $\mathit{D}_d$, containing $d$ possible target tokens including the whole valid vocabulary tokens and one blank token $-$. 

Let $\hat z_i(x;q)$ denote the ``path'' from the query set to the output token, meaning the value of $z$ as a (deterministic) function of the encoding $x$ for a given set of LQTs $q$.  We denote the ensemble of paths $p(z | x; q) = \delta (z- \hat z(x;q))$, where $\delta$ is Dirac function and $q$ are learnable parameters. 
The ordinary cross-entropy loss would be $
L_{CE}(\theta) = -  \sum_{i} \text{log} \frac{e^{f_{y_i}(z_i) }}{\sum_j e^{f_{y_j}(z_i)}}
$
where $f_{y_j}(\cdot)$ computes the logits of $x_i$ on target token class $y_i$, and $\theta$ are the learnable parameters of the encoder that produces $x$, and of the decoder that produces $z$ given $x$.

Given the vocabulary size $d$ and output sequence length $N$, there are $N^{d}$ possible output sequences. Among all output sequences, the Q-CTC selects the valid output sequences, and maximizes the probability over all valid paths. Collapse operation, denoted as $\mathcal{B}$, removes all blank tokens $-$, and merge continuous repetitive tokens between two blank tokens. For example, $\mathcal{B}(\text{$-$ a bag on a table})=\mathcal{B}(\text{a $-$ bag bag $-$ on a a table $-$})=\text{a bag on a table}$. There are many possible sequences that can be collapsed to the correct target sequence, which are valid alignments. When we calculate the loss, we marginalize over the set of valid paths.  
We denote the collapsed valid sequences as $\tilde{p}(z | x; q)$. 

Before $x$ is instantiated at inference time, $z_i$ are random variables, functions of $x$, which we need to marginalize in the loss, obtaining 
\begin{equation}
L_{\rm \text{Q-CTC}}(\theta, q) = -  \sum_{k}  \sum_{z_i \sim 
\tilde{p}(z|x_k; q)} \text{log}\frac{e^{f_{y_i}(z_i(x_k; q))}}{\sum_j e^{f_{y_j}(z_i(x_k; q))}}
\label{eq:LQC}
\end{equation},
where $k$ is the number of all training samples.
Note that this loss is different from the CTC loss, where the marginalization only depends on $x$, not $q$. Our loss is a function of $\theta$ {\em as well as the learnable parameters} used to produce $q$, in addition to any other hyperparameters shared with the ordinary autoregressive model.

\subsubsection{Knowledge Distillation}
Optimising Q-CTC is more challenging in visual-language domain than language domain, because the image-text input sequence has less structure than pure language inputs, and the model is required to generate well-structured sequences from less order sequences.    
Due to the large solution space and the lack of inter-token dependency between decoder tokens, non-autoregressive models can have lower performance compared to auto-regressive models. To reduce the solution space in model training, we exploit two simple knowledge distillation mechanisms\cite{hinton2015distilling} in training. Specifically, we train a standard Transformer with an auto-regressive loss, and set this model as the teacher model, which has significantly less variations in ground-truth and makes learning targets more deterministic. We observe that this knowledge distillation benefits the task of long sequence outputs such as image captioning, and the detailed discussion is in Section \ref{sec:benchmark}. We also observe that the model initialization weights are non-trivial for training Q-CTC loss, and we use OFA pretrained weights to initialize NARVL and finetune it in various down-stream tasks. 
 \vspace{-4mm}

\subsubsection{Model inference}
\vspace{-2mm}
During the inference, for each output token, we use the text token in the vocabulary with the largest probability as prediction. We follow the same path collapse rules used in training to remove - and repetitive tokens to obtain the final predicted sequence.

\begin{figure}[!h]
\includegraphics[width=0.48\textwidth]{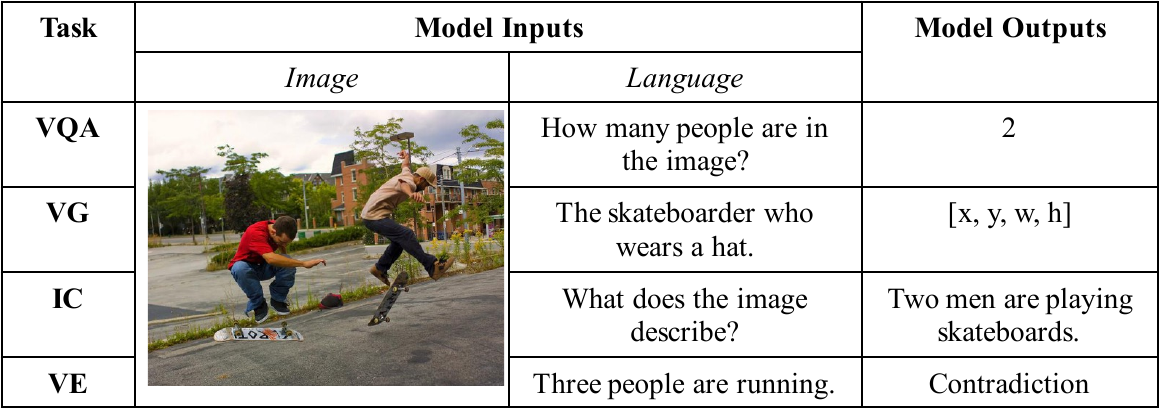}
\caption{We test the proposed NARVL on various vison-language tasks, including Visual Question Answering (VQA), Visual grounding (VG), Image Captioning (IC) and Visual Entailment (VE). The inputs and outputs of each tasks are illustrated here, and all types outputs are unified within the sequence formulation.
}
\vspace{-1mm}
\label{fig:task}
\end{figure}

\vspace{-2mm}

\section{Experiments}
\label{sec:experiments}

We perform experiments on various vision-language tasks, and make performance and speed comparisons to the state-of-the-art models to show the effectiveness of the proposed method. Figure \ref{fig:task} summarizes the tasks we experiment on. 

\subsection{Implementation details}
We implement NARVL starting from the official OFA\cite{wang2022ofa} code, which is written in the fairseq library \cite{ott2019fairseq}. To make a fair comparison to the autoregressive OFA model, we keep most hyper-parameters and training schedule the same and this helps us to understand the effect of switching decoder paradigm. Some OFA models are reproduced to get the weights for speed benchmarking purpose, and are noted in the result tables. We use the same OFA encoder task prompts, but don't use decoder prompts, as the NARVL decoder doesn't do conditional token generation. 

Q-CTC loss is used in the final model of all tasks, and knowledge distillation is only used in Image Captioning and VQA tasks. We follow the model size settings used in \cite{wang2022ofa}.
We benchmark our model speed on Tesla V100-SXM2-16GB GPU with a batch size of 1, and take the average over the test samples. As the GPU needs to ramp up, the first image tends to be much slower than other images, and we remove the first sample inference time in all calculations. Gradient accumulation is used at training time to increase the batch size of the below experiments. 

\noindent{\bf Referring Expression Grounding} 
RefCOCO\cite{yu2016modeling}, RefCOCO+\cite{yu2016modeling}, RefCOCOg\cite{mao2016generation} datasets are created based from the COCO dataset, where a piece of text (referring expression) that describes a unique object in the image is given and the model is asked to find the object. 

In both RefCOCO and RefCOCO+, testA only contains people and testB only contains non-people objects. 
We feed the referring expression text sequence and image to the encoder, and let the decoder predict the fixed length output sequence of bounding box position tokens. 

When benchmarking the model speed, we take the average inference time on the entire validation or test set. On RefCOCO and RefCOCOplus, we average the inference time of val, testA and test B subsets and only report the average inference time for simplicity. Similarly we average over val and test subsets for RefCOCOg dataset. We train NARVL for 10 epochs with effective batch size of 128, and we set the number of decoder learnable query input tokens to 5. 

\noindent{\bf Visual Entailment}
SNLI-VE dataset is built off SNLI and Flickr30K datasets, and the task requires the model to reason relationship (entailment, contradiction and neutral) between an image premise and text sentence hypothesis. Image premise and text sequence hypothesis are fed into the NARVL encoder, and the decoder predicts the sequence of one token that is one of the relationship words. We use a batch size of 256 and train our model for 5 epochs, and the number of decoder input learnable query tokens is set to 2. Knowledge distillation is not adopted given the simplicity of the output sequence of this task.

\noindent{\bf Visual Question Answering}
Visual Question Answering is a task that requires cross-modal reasoning that the model is asked to answer question by looking at image. The number of decoder input tokens is set to 6, and the model is trained with 10 epochs with a batch size of 512. We don't use any candidate answer set based constraint on the generated answer sequence. We found the default OFA model that uses all-candidate inference is quite slow, and we additionally benchmark and report the faster beam-search version that was released in the OFA github repository. Speed is benchmarked on the validation set, and the accuracy numbers are obtained from the official evaluation server. 

\noindent{\bf MSCOCO Image Captioning}
Image captioning requires models to generate a fluent and meaningful natural language sentence that describes the image. We use batch size of 128 and train 5 epochs with decoder input sequence length of 20. 
Each image in the caption dataset has 5 captions written independently from 5 annotators, which increases the difficulty for NARVL training which might merge possible captions and generate captions that are not fluent, and knowledge distillation is adopted to simplify the training captions.

\vspace{-1mm}
\subsection{NAR vs AR}
\vspace{-1mm}
We compare accuracy and speed between AR (autoregressive) and our proposed NAR (non-autoregressive) models on various vision-
language tasks and results are shown in Table \ref{tab:nar_vs_ar}. It shows that the NAR model consistently outperforms the AR model on visual grounding and has significantly higher execution
speed (with 2.4 to 12.7 times speed up on various tasks). As visual entailment requires shorter output sequences, the speedup is smaller than for other tasks. The significant speedups are
on VQA and Captioning datasets, because the length of output sequence is longer than that of the grounding task. We will analyse the accuracy and inference speed for each downstream tasks in next section.

\begin{table*}[th]

\centering
 \caption{Accuracy and speed comparisons of AR (autoregressive) and our proposed NAR (non-autoregressive) models on various vision-language tasks. Our NAR model is trained under the exact same settings of model size, parameters and training schedule as the AR model for fair comparison. The NAR model consistently outperforms the AR model on visual grounding and has significantly higher execution speed. As visual entailment requires shorter output sequences, the speedup is smaller than for other tasks. We observe significant speedups on VQA and Captioning datasets, but with a measurable performance drop. Beam search is used in all AR models and greedy decoding is used in NAR models (beam search can be applied and are studied in \ref{table:albation}, but not adopted due to its sequential nature.). All models reported here are in base size. Inference wall clock time is measured in ms.}
\scalebox{0.86}{
\begin{tabular}{c|ccccccccccccc}
\hline
&\multicolumn{10}{c}{ (a) Visual Grounding }\\ \hline
\rule{0pt}{10pt}&\multicolumn{4}{c|}{ RefCOCO } &\multicolumn{4}{c|}{{RefCOCO+}}&\multicolumn{3}{c}{{RefCOCOg}}\\ 
\rule{0pt}{10pt}{Method}& {Val} & {TestA} & {TestB}  & \multicolumn{1}{c|}{Time}& {Val} & {TestA} & {TestB}  & \multicolumn{1}{c|}{Time}& {Val} & {Test} & {Time}\\
\rule{0pt}{10pt}AR & 88.15 &  90.08  & 83.45 & \multicolumn{1}{c|}{133.6/1$\times$}&   81.67 &  86.40 & 74.49 & \multicolumn{1}{c|}{133.1/1$\times$}&  81.92 &  82.02 &  132.8/1$\times$ \\
\rowcolor{gray!15} \rule{0pt}{10pt}NAR & \textbf{88.78} & \textbf{90.63} & \textbf{84.67} & \multicolumn{1}{c|}{\textbf{54.8 / 2.4$\times$}}& \textbf{82.35}  & \textbf{87.15} & \textbf{74.74} & \multicolumn{1}{c|}{\textbf{54.6/2.4$\times$}}& \textbf{82.27} & \textbf{82.69} &   \textbf{54.8/2.4$\times$} \\\hline

\rule{0pt}{10pt}&\multicolumn{3}{c|}{ (b) Visual Entailment }&\multicolumn{3}{c|}{ (c) Visual Question Answer }&\multicolumn{5}{c}{ (d) Image Captioning }\\  
\rule{0pt}{10pt} {Method}& {Dev} & {Test} & \multicolumn{1}{c|}{Time}& {Test-dev} & {Test-std} & \multicolumn{1}{c|}{Time}& {BLEU@4} & {METEOR} & {CIDEr} & {SPICE} & {Time}\\
 \rule{0pt}{10pt}AR& {89.0} & {89.0} & \multicolumn{1}{c|}{68.0/1$\times$}&\textbf{77.48} & \textbf{77.58} & \multicolumn{1}{c|}{645.1/1$\times$}& \textbf{41.0} & \textbf{30.9} & \textbf{138.2} & \textbf{24.2} & 366.0/1$\times$\\
\rowcolor{gray!15} \rule{0pt}{10pt}NAR& 89.0 & 89.0 & \multicolumn{1}{c|}{\textbf{48.7/1.4$\times$}}& 75.69 & 75.75 & \multicolumn{1}{c|}{\textbf{50.7/12.7$\times$}} & 36.4 & 28.7 & 123.1 & 22.5 & \textbf{51.2/7.2$\times$}\\
 
\hline
\end{tabular}
}
 \label{tab:nar_vs_ar}
\end{table*}

\subsection{Benchmark Performance}
\label{sec:benchmark}

\begin{table}[!th]
\centering
\caption{Results on visual grouding datasets: RefCOCO, RefCOCO+ and RefCOCOg, and  comparisons to previous works. * Weights are not released and the model was reproduced by us using the official released training scripts. Difference to the reported results in \cite{wang2022ofa} might due to randomness in checkpoint picking. We do not perform knowledge distillation for this task.}
% \vspace{-2mm}
\label{table:refcoco}
\scalebox{0.8}{
\begin{tabular}{cccccc}
\hline
\multicolumn{6}{c}{{RefCOCO}} \\
{Method}& {Val} & {TestA} & {TestB}  & {Speed (ms)} &\\\hline
UNITER\cite{chen2020uniter} & 81.41 &  87.04 &  74.17 & - &\\
VILLA\cite{gan2020large} &  82.39 &  87.48 &  74.84 & - &\\
MDETR\cite{kamath2021mdetr} & 86.75 & 89.58 &   81.41 & 73.5 &\\
UNICORN\cite{yang2021crossing} &  88.29 &   90.42 &  83.06 & 266.4 &\\
OFA*\cite{wang2022ofa} & 91.24 &  93.41  & 87.16 & 284.9 &\\\hline

\rowcolor{gray!15} NARVL$_{tiny}$(ours) & 80.40 & 84.64 & 73.8 &  \textbf{30.7} &\\
\rowcolor{gray!15} NARVL$_{base}$(ours) & 88.78 & 90.63 & 84.67 & 54.9 &\\
\rowcolor{gray!15} NARVL$_{huge}$(ours) & \textbf{91.8} & \textbf{94.24} & \textbf{88.01} & 150.8 &\\

\hline\hline
\multicolumn{4}{r}{RefCOCO+} & \multicolumn{2}{r}{RefCOCOg} \\
{Method} & {Val} & {TestA} & {TestB} & {Val} & {Test} \\\hline

UNITER\cite{chen2020uniter} &  75.90 &   81.45 &  66.70 &  74.86 & 75.77  \\
VILLA\cite{gan2020large} &   76.17 &    81.54 &  66.84 &  76.18 & 76.71  \\
MDETR\cite{kamath2021mdetr} & 79.52 &  84.09 &  70.62 &  81.64 &  80.89  \\
UNICORN\cite{yang2021crossing} &  80.30 &   85.05 &   71.88 &  83.44 & 83.93 \\
OFA* \cite{wang2022ofa} &   86.93 & 91.37 & 80.51 &  86.38 &  87.70  \\\hline

\rowcolor{gray!15} NARVL$_{base}$(ours) & 82.35 & 87.15 & 74.74  & 82.27 & 82.69 \\
\rowcolor{gray!15} NARVL$_{huge}$(ours) & \textbf{87.90}  & \textbf{92.18} & \textbf{81.2} & \textbf{87.7} & \textbf{88.42}  \\

\hline
\end{tabular}
}
\vspace{-1mm}
\end{table}
\noindent{\bf RefCOCO, RefCOCO+, RefCOCOg results.} Following the metric used previous works, we report Acc@0.5 numbers.  We compare NARVL to other methods in Table \ref{table:refcoco}. Our proposed NARVL model achieves state-of-the art performance on all subsets of RefCOCO, RefCOCO+ and RefCOCOg datasets. As shown in Table \ref{tab:nar_vs_ar}, our NAR model consistently outperforms the AR model on all subsets of the three datasets and has significantly faster speed. Take RefCOCO dataset as an example, our NAR model has on average 0.83 higher accuracy compared to the AR model, and is 2.4 times faster (54.8 ms vs 133.6 ms). The speedup is introduced by the parallel nature of NARVL decoder, and we argue the accuracy improvements come from the bidirectional attention of our parallel decoder, as opposed to uni-directional attention in autoregressive decoder, where the later generated coordinate tokens are not available in attention operation of the early token generation.

\begin{table}[h]
\centering
\caption{{SNLI-VE visual entailment results. NARVL shows on par performance to previous state-of-the-art. Knowledge distillation is not used in this task.}}
% \vspace{-2mm}
\label{table:visual_entailment_results}
\scalebox{0.9}{
\begin{tabular}{@{}lcccccccccc@{}}
\hline
{Method} & {dev} & {test}
\\\hline
UNITER\cite{chen2020uniter} &  73.8 &  74.0\\
VinVL\cite{zhang2021vinvl} &  76.5 &   76.6 \\
UNIMO\cite{li2020unimo} &  75.0 &   75.3 \\
ALBEF\cite{li2021align} &  75.8 &   76.0 \\
METER\cite{dou2022empirical} &  77.7 &   77.6 \\
VLMo\cite{wang2021vlmo} &  79.9 &   80.0 \\
SimVLM\cite{wang2021simvlm} &   80.0 &    80.3\\
Florence\cite{yuan2021florence} &   80.2 &  80.4 \\
% OFA$_{base}*$\cite{wang2022ofa} & 89.3 & 89.2 & 68.0 \\
OFA \cite{wang2022ofa} & 91.0 & \textbf{91.2} &\\\hline
\rowcolor{gray!15} NARVL$_{base}$(ours) & 89.0 & 89.0\\
\rowcolor{gray!15} NARVL$_{huge}$(ours) & \textbf{91.1} & 91.1 \\
 
\hline
\end{tabular}
}
 
\end{table}

\noindent{\bf SNLI-VE results} We compare our NAR model and AR model in Table \ref{table:visual_entailment_results}. Our model has average inference time of 48.73 ms, which is 1.4 times faster than the AR model with 68.03 ms. The effective target sequence excluding EOS has only 1 token, which means decoder attention in essence is the same for NAR and AR model, and we see exactly the same performance. As the output sequence (label-EOS) for this dataset is shorter than other dataest, the relative speedup from NAR model is smaller. Our NARVL shows on par performance as the current state-of-art OFA model on both validation and test set, as shown in Table \ref{table:visual_entailment_results} along with comparisons to other previous methods.

\noindent{\bf VQA results} Comparisons to the AR model on VQA are shown in \ref{tab:nar_vs_ar}. NAR model performance is on average 1.81 lower than the AR model, while it has gigantic speed up of 12.7 times (50.7 ms vs 645.1 ms). Our model shows competitive results and more comparisons to previous methods can be found in Table \ref{table:vqa_results}. The default OFA model uses a slow all-candidate decoding method, and we additionally report the results with beam search decoding. It is noteworthy that OFA-VQA was trained with trie-based auto-regressive decoding that predicts next node on a Trie in each step, and the speedup would be smaller over standard AR models. In this task, one question might have multiple valid answers e.g. "What's the color of the shirt?" can be answered with "Red and White" or "White and Red".This might create confusions for NARVL, which leads to incorrect predictions like "White and White", and here we use knowledge distillation from an AR model of the same size to make the answers used in training more deterministic.

\begin{table}[!h]
\centering
\caption{{Results on VQA.} We report base version of NARVL with knowledge distillation. *beam search version is released in the official OFA github that is much faster than the original allcnd version reported in the OFA paper, and we present results from both models here.}
% \vspace{-2mm}
\label{table:vqa_results}
\setlength{\tabcolsep}{3.0pt}
\scalebox{0.8}{
\begin{tabular}{@{}lcccccccccc@{}}
\hline
{Method} & {test-dev} & {test-std} & {Speed (ms)} \\\hline
UNITER\cite{chen2020uniter} &  73.8 &  74.0 & - \\
UNIMO\cite{li2020unimo} &  75.0 &   75.3 & - \\
ALBEF\cite{li2021align} &  75.8 &   76.0 & - \\
METER\cite{dou2022empirical} &  77.7 &   77.6 & - \\
VLMo\cite{wang2021vlmo} &  79.9 &   80.0 & - \\
SimVLM\cite{wang2021simvlm} &   80.0 &    80.3 & - \\
Florence\cite{yuan2021florence} &   80.2 &  80.4 & - \\
OFA$_{huge}$ Allcan\cite{wang2022ofa} & \textbf{82.0} & \textbf{82.0} & - \\
OFA$_{base}$ Beam Search*\cite{wang2022ofa} & 77.48 & 77.58 & 645.1 \\
OFA$_{base}$ Allcan\cite{wang2022ofa} & 78.0 & 78.1 & 15415.3 \\\hline
\rowcolor{gray!15}NARVL-KD$_{base}$ (ours) & 75.59 & 75.75 & \textbf{50.76}\\
\rowcolor{gray!15}NARVL-KD$_{huge}$ (ours) & 79.59 & 79.39 & 81.53\\
\hline
\end{tabular}
}
% \vspace{-2mm}
\end{table}

\noindent{\bf MSCOCO Image Captioning results.} Comparisons in Table \ref{table:distill} demonstrate the superiority of the Q-CTC loss function in the context of NARVL, where we observe a huge improvement on all metrics with Q-CTC loss. On top of the Q-CTC version of model, we utilize knowledge distillation from AR model and obtained another set of massive performance boost, as shown in Tab \ref{table:distill}.  Comparison results of our NAR model to the AR model are shown in Tab \ref{tab:nar_vs_ar}, and we see huge speed advantage of the NAR model and performance advantage of the AR model. More comparisons to previous methods are shown in Table \ref{table:coco_caption}, where all models shown are trained without CIDEr reinforcement learning optimization, and NARVL shows competitive results.

\begin{table}[h]
\centering
\caption{{Results on MSCOCO Image Captioning Karpathy test split. All the models are trained without CIDER reinforcement learning.}}
% \vspace{-2mm}
\label{table:coco_caption}
\scalebox{0.68}{
\begin{tabular}{@{}lccccccc@{}}
\hline
{Method} & {BLEU@4} & {METEOR} & {CIDEr} & {SPICE} & {Speed (ms)} \\
\midrule
VL-T5\cite{cho2021unifying}  &  34.5 & 28.7 & 116.5 & 21.9 & - \\
OSCAR\cite{li2020oscar} &  37.4 & 30.7 & 127.8 & 23.5 & - \\
UNICORN\cite{yang2021crossing} &  35.8 & 28.4 & 119.1 & 21.5 & - \\
VinVL\cite{zhang2021vinvl} & 38.5 & 30.4 & 130.8 & 23.4 & - \\
LEMON\cite{hu2022scaling}  & 41.5 & 30.8 & 139.1 & 24.1 & - \\
SimVLM\cite{wang2021simvlm}  & 40.6 & \textbf{33.7} & 143.3 & \textbf{25.4} & - \\
OFA\cite{wang2022ofa}  & \textbf{43.9} & 31.8 & \textbf{145.3} & 24.8 & 545.1 \\
\hline
\rowcolor{gray!15} NARVL-KD$_{base}$ (ours) & 36.4 & 28.7 & 123.1 & 22.47 & \textbf{51.2} \\
\rowcolor{gray!15} NARVL-KD$_{huge}$ (ours) & 40.1 & 30.6 & 136.7 & 24.3 & 130.6 \\
\hline
\end{tabular}
}
\end{table}

\section{Ablation experiments} 
\label{sec:ablation}
 
{\noindent\bf{Sequential learnable queries in decoder}}
We proposed the learnable query token as the decoder input, which is shown in Figure \ref{fig:nar}. This design differs from the existing non-autoregressive Transformer for sequence generation, which uses the outputs of the encoder as the inputs for the decoder \cite{gu2020fully}. We compare these two designs on RefCOCO dataset and MSCOCO Image Captioning dataset, and the results are shown in Table \ref{table:decoder_design_refcoco} and Table \ref{table:decoder_design_caption}, respectively. We observe that our learnable query token design has both accuracy and speed advantage over the encoder output design. 

\begin{table}[th]
\centering
\caption{Comparisons of two types of decoder input design on RefCOCO dataset. The learnble query token design is faster and more accurate.}
\label{table:decoder_design_refcoco}

\scalebox{0.8}{
\begin{tabular}{ccccc}
\hline
{Decoder Input}& {Val} & {TestA} & {TestB}  & {Speed (ms)}\\\hline
Output of Encoder & 87.78 & 89.89 & 83.24 & 82.8 \\
Learnable Query Tokens(ours) & \textbf{88.78} & \textbf{90.63} & \textbf{84.67} & \textbf{54.9} \\
 
\hline
\end{tabular}
}
\vspace{-0mm}
\end{table}

\begin{table}[th]
\centering
\caption{Comparisons of two types of decoder input design on MSCOCO Image Captioning dataset. The learnble query token design has equal or better performance with faster inference speed.}
\vspace{-2mm}
\label{table:decoder_design_caption}
\scalebox{0.65}{
\begin{tabular}{@{}lcccccccc@{}}
\hline
{Decoder Input} & {BLEU@4} & {METEOR} & {CIDEr} & {SPICE} & {Speed (ms)} \\
\midrule
Output of Encoder & 36.4 & 28.6 & 121.8 & 22.0 & 58.7 \\
Learnable Query Tokens(ours) & 36.4 & \textbf{28.7} & \textbf{123.1} & \textbf{22.5} & \textbf{51.2} \\
\hline
\end{tabular}
}
\vspace{-0mm}
\end{table}

{\noindent\bf{Q-CTC loss vs Cross-entropy loss}} We compare Q-CTC loss and standard cross-entropy loss (CE) in NARVL, and the results are in Table \ref{table:ctc_vs_ce}. With cross entropy loss, the first $n$ input queries are supervised to predict the output sequence, where $n$ is the number of tokens in the target sequence. Q-CTC loss performs significantly better than CE loss as it assigns proper penalty to the model, while CE penalizes the model severely even just one token position shift.

\begin{table}[!h]
\centering
\caption{{Comparing Q-CTC loss and CE loss with NARVL on Captioning. The base model is used in the experiments and knowledge distillation is not used for both losses.}}
% \vspace{-2mm}
\label{table:ctc_vs_ce}
\scalebox{0.8}{
\begin{tabular}{@{}lcccccccc@{}}
\hline
% \multicolumn{7}{c}{\textbf{MSCOCO Image Captioning}} \\
{Loss} & {BLEU@4} & {METEOR} & {CIDEr} & {SPICE} & {Speed (ms)} \\
\midrule
CE & 17.54 & 18.57 & 66.85 & 11.89 & \textbf{52.05} \\
Q-CTC & \textbf{26.69} & \textbf{24.06} & \textbf{93.24} & \textbf{17.38} & 54.03 \\
Difference  & +9.15 & +5.49 & +26.39 & +5.49 & +2.0 \\
 
\hline
\end{tabular}
}

\vspace{-0mm}
\end{table}
% \vspace{-0.4cm}

{\noindent\bf{Knowledge distillation}}
We study the effect of knowledge distillation on VQA datasets, and found it improves the performance by 1.48/1.35 on test-dev/test-std splits, as shown in Tab \ref{table:distill}. 
\begin{table}[!h]
\caption{{Effect of knowledge distillation on VQA and Captioning datasets. The models with "KD" are the distilled models.} }
% \vspace{-2mm}
\label{table:distill}

\begin{subtable}[h]{0.45\textwidth}
\centering
\scalebox{0.8}{
\begin{tabular}{cccccc}
\toprule
{Method} & {BLEU@4} & {METEOR} & {CIDEr} & {SPICE} & {Speed (ms)} \\
\midrule
NARVL & 26.7 & 24.1 & 93.2 & 17.4 & 54.0 \\
NARVL-KD & \textbf{36.4} & \textbf{28.7} & \textbf{123.1} & \textbf{22.5} & \textbf{51.2} \\
Difference & +9.7 & +4.6 & +29.9 & +5.1 & -2.8 \\
\hline
\end{tabular}
}

\vspace{2mm}
\caption{MSCOCO Image Captioning}
\end{subtable}
\vspace{2mm}
\vfill
\begin{subtable}[h]{0.45\textwidth}
\centering
\scalebox{0.85}{
\begin{tabular}{cccccc}
\toprule

{Method} & {Test-dev} & {Test-std}  & {Speed (ms)} \\\hline
NARVL & 74.21 & 74.4 &  51.52 \\
NARVL-KD & \textbf{75.69} & \textbf{75.75} &  \textbf{50.76} \\
Difference & +1.48 & +1.35 & -0.76 \\
\bottomrule
\end{tabular}
}
\caption{VQA}
\end{subtable}

 \vspace{-3mm}
\end{table}

{\noindent\bf{The various lengths of query tokens}} We study the effect on performance and speed of our method when we change the number of input queries to the decoder and decoding methods. Ablation experiments are done on MSCOCO Image Captioning dataset, and the results are shown in Table \ref{table:albation}. If the input length is smaller than the target sequence, the decoder won't be able to generate complete sequence, which leads to reduced performance for number of values that are too small, while too large number of input queries makes it harder for the model to decide the input output token correspondence. Naturally larger number of queries leads to increased inference time, while increasing the number of input queries from 10 to 1000 only leads to an increase from 51.32 ms to 60.40 ms, demonstrating the amazing ability of long sequence scalability. 

\begin{table}[th]
\centering
\caption{{Ablation experiments for the number of queries on Image Captioning dataset. All experiments are done with base size model. The models used in query number ablations only and 1 epoch training)}}
\vspace{-0mm}
\label{table:albation}
\scalebox{0.7}{
\begin{tabular}{@{}lcccccccc@{}}
\hline
\multicolumn{6}{c}{Decoder sequence length ablations} \\
Number of Queries & {BLEU@4} & {METEOR} & {CIDEr} & {SPICE} & {Speed (ms)} \\
\midrule
10 & 29.43 & 24.89 & 96.90 & 18.51 & 51.32 \\
20 & 33.18 & 27.10 & 110.35 & 20.48 & 51.72 \\
100 & 32.47 & 26.98 & 108.76 & 20.42 & 51.96 \\
500 & 32.36 & 26.79 & 107.34 & 19.94 & 53.82 \\
1000 & 31.91 & 26.60 & 105.90 & 19.89 & 60.40 \\
\hline
\end{tabular}
}
\vspace{-0mm}
\end{table}

{\noindent\bf{Beam search in NARVL}}
We also experiment with greedy and beam search decoding, and observe slightly better performance from beam search. Due to the increased inference time and autoregressive nature of beam search decoding, it's not adopted in previous experiments. 
 
\begin{table}[th]
\centering
\caption{{Ablation experiments of Beam Search in NARVL. Test it on Image Captioning dataset. All experiments are done with base size model.}}
\vspace{-1mm}
\label{table:beam}
\scalebox{0.75}{
\begin{tabular}{@{}lcccccccc@{}}
\hline
\multicolumn{6}{c}{Decoding method experiments} \\
{Decoding Method} & {BLEU@4} & {METEOR} & {CIDEr} & {SPICE} & {Speed (ms)} \\
\midrule
Greedy & 36.38 & 28.70 & 123.15 & 22.46 & 51.18 \\
Beam Search 5 & 36.85 & 28.74 & 124.03 & 22.51 & 68.95 \\
Beam Search 20 & 36.91 & 28.71 & 124.15 & 22.51 & 70.22 \\
Beam Search 100 & 37.00 & 28.71 & 124.25 & 22.51 & 122.91 \\
\hline
\end{tabular}
}
\vspace{-0mm}
\end{table}

\begin{figure}[!h]
%\vspace{-15mm} 
\begin{center}
\begin{tabular} {c}
\includegraphics[width=0.30\textwidth]{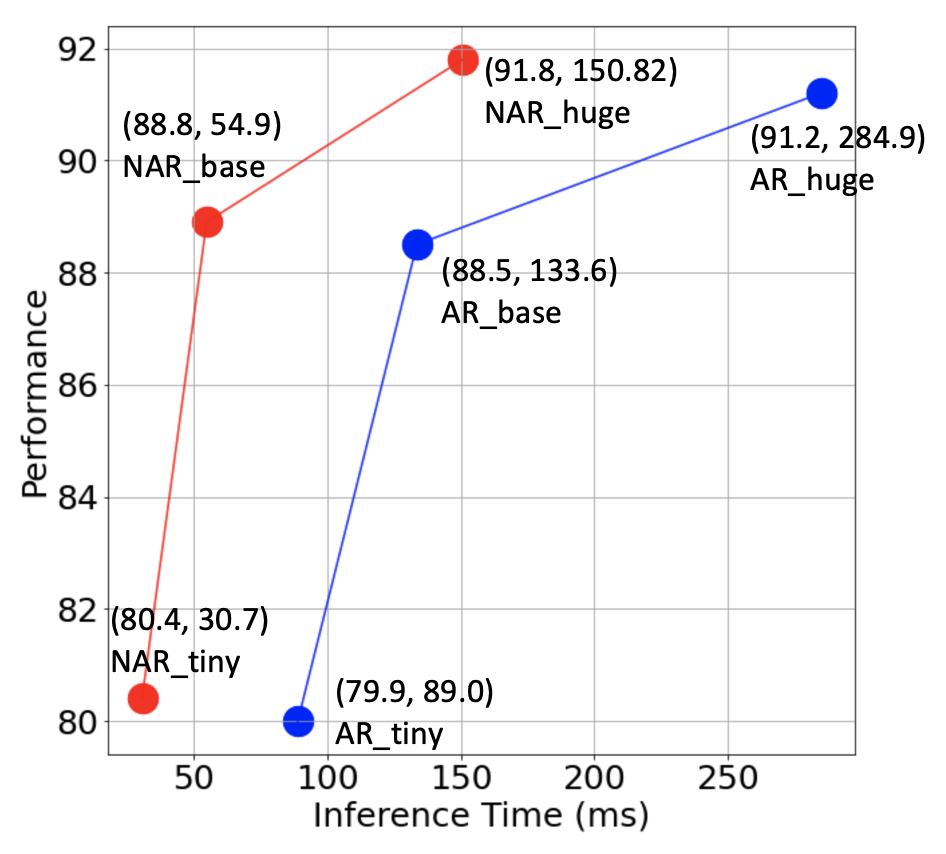}
\end{tabular}
\end{center}
\vspace{-5mm}
\caption{The comparison of accuracy and inference speed with NAR (non-autoregressive) and AR (autoregressive) models for varying model sizes: Tiny, Base and Huge. Speed is measured in wall clock time. NAR models significantly outperform their AR counterparts  under the same inference time budget on the RefCOCO validation set.
}
\vspace{-5mm}
\label{fig:size}
\end{figure}

{\noindent\bf{The model scale}}
We investigate the performance and inference speed of AR and NAR models of different model sizes, and the comparison is illustrated in Figure \ref{fig:size}. We observe that the inference latency increases when the model becomes large on both AR and NAR model, but AR model has larger latency change compared to AR models. It is worthy to note that NAR\_$\text{huge}$ has the similar the similar performance as AR\_$\text{huge}$, meanwhile its inference speed is closed to the AR\_$\text{base}$ model. 

\section{Limitations}
Optimization of the NAR models is difficult, and the best performance we get on Image Captioning and VQA datasets rely on distillation from an autoregressive model, which is inconvenient in practice as one needs to train two models. The most common failure case comes from the conditional  independence nature between the decoder tokens, and the model might end up merging multiple possible valid sequences into a incorrect output sequence.

\section{Conclusion}

We have introduced NARVL, an All-in-One non-autoregressive model for various visual-language tasks, and the proposed NARVL repurposes the autoregressive decoder into a more flexible encoding that can be tailored to different hypothesis spaces using a layer of learnable query tokens (LQTs). These tokens are, in turn, used to define Query-CTC loss, akin to losses used in language modeling, but augmented to incorporate LQTs. This innovation is key to enabling the flexibility of letting the task drive the design of the hypothesis space, which we deem critical for heterogeneous input and output spaces as expected in the visual domain but absent in language.

\newpage
%%%%%%%%% REFERENCES
{
    \small
    \bibliographystyle{ieeenat_fullname}
    \bibliography{main}
}

% WARNING: do not forget to delete the supplementary pages from your submission 

% \appendix
% \input{X_suppl}

\end{document}